# Real-time Load Prediction with High Velocity Smart Home Data Stream

**Christoph Doblander · Martin Strohbach · Holger Ziekow · Hans-Arno Jacobsen**

**Abstract** This paper addresses the use of smart-home sensor streams for continuous prediction of energy loads of individual households which participate as an agent in local markets. We introduces a new device level energy consumption dataset recorded over three years wich includes high resolution energy measurements from electrical devices collected within a pilot program. Using data from that pilot, we analyze the applicability of various machine learning mechanisms for continuous load prediction. Specifically, we address short-term load prediction that is required for load balancing in electrical micro-grids. We report on the prediction performance and the computational requirements of a broad range of prediction mechanisms. Furthermore we present an architecture and experimenal evaluation when this prediction is applied in the stream.

**Keywords** Electrical load prediction · Machine learning

## 1 Introduction

In recent years a range of technologies emerged that brings the Internet of Things (IoT) into the domain of buildings. The combination of sensors and intelligent analytics in households is referred to as smart home and promises to enable a wide range of applications. In

C. Doblander
E-mail: doblande@in.tum.de

M. Strohbach
E-mail: MStrohbach@agtinternational.com

H. Ziekow
E-mail: HZiekow@agtinternational.com

H.-A. Jacobsen
E-mail: jacobsen@in.tum.de

particular, utilities see potential in better analyzing and managing energy loads [7, 12, 13, 15]. Furthermore, application scenarios extend to solutions for comfort [27], safety and security [18], and care of the elderly [23]. Utilities see huge potential in smart home technology to better balance electrical loads within their supply grids [5,25]. Keeping production and consumption equal at all times is critical for grid stability and the key responsibility of grid operators. Oversupply as well as undersupply lead to reduced power quality, malfunctioning or even damage of electrical devices, and potentially to the collapse of grid. Balancing out the delta between planned supply and actual energy usage requires costly investments in technologies that can compensate the delta in real-time (e.g., batteries). The increasing portion of renewable energy elevates these challenges to a new level [6]. With renewable energy, power is no longer fed into the grid from centrally controlled big power plants, but supplied by many small and distributed generators. The consequence is that balancing supply and demand needs to happen at a local level as well, i.e., at the so called micro grid. In this situation utilities are left with two options. One option is to invest in physical infrastructure to better cope with imbalances in the micro grid (e.g., additional transmission lines and storage solutions). The other complementary option is to enhance energy analytics [9] for better managing supply and demand on a local grid level, i.e., within the distribution grid or neighborhood. Predicting the consumption is a prerequisite for planning the supply such that it matches the demand. But not only accurate predictions are relevant but also knowing the prediction error is important. Reducing the error also reduces the need for stand-by equipment to balance short-term inbalances.

In our experiment, we send sensor information from the smart home controller to a analytics pipeline. The pipeline predicts the energy consumption for multiple time horizons in future. This predictions can be used by an agent that trades on the local energy market within the grid constraints. Further the prediction error can be used to provision controlable short term resources or demand response (DR) mechanisms. Such ideas are piloted in several projects [4,16,29]. The main idea is that individual consumers or devices adapt their consumption based on signals in order to better fit the supply. For instance, an AC may temporally power down to compensate for a drop in supply [2]. However, like supply planning, demand response needs a prediction of the demand to imitate appropriate actions for demand control. The need to balance out production on the level of neighborhoods shifts the requirements for consumption prediction to a new level. Instead of predicting the consumption for cities or regions, the scope narrows to the level of individual households or even individual devices (e.g., to initiate a demand response signal). Smart home solutions provide the data on the required level of detail but call for analytics solutions that can cope with the corresponding data volumes. For instance, in this paper, we report on a pilot smart home system that streamed about 1.2 million data records per house and day. Furthermore, the streaming nature of smart home data and the need for real-time actions in grid management call for analytics solutions that enable real-time prediction of high volume data streams.

In this paper, we contribute to the development of real-time forecasting solutions with smart home data. Our main contributions are the following:

1. We introduce a high resolution smart home data set that was captured in a pilot project over several years.
2. We analyze the prediction performance of a broad range of machine learning mechanisms for short-term load forecasting with smart home data.
3. We present a scalable solution for training household specific forecasting models and continuous prediction within smart home data streams.

The rest of the paper is structured as follows: In Section 2, we give an introduction to the project. Then, in Section 3, we introduce the dataset. In Section 4, we detail our approach of prediction. In Section 5, we show our proposed architecture for prediction of streams and batch analytics. In Section 6, we present the results in terms of prediction error and computational load. In Section 7 we discuss the relevant related work and we conclude in Section 8.

## 2 Project Background

The evaluations of the prediction algorithms are based on smart home data that was recorded as part of the PeerEnergyCloud project [3]. The project was funded by the German Federal Ministry for Economics Affairs and Energy and addressed technologies for improving load balancing in local energy grids. It was carried out by a consortium of research institutions and industry, including the local utility company in the German city Saarlouis. Part of the project was a field trial of a smart home system with several dozen houses, which was conducted with the utility company in Saarlouis. Participating houses received a set of smart home sensors for data collection, providing the basis for research on data analysis in the project. The pilot started in 2011 and is in part still running.

One goal of the project was the investigation of means to foster use of locally produced energy (e.g., from solar panels of private houses) in close vicinity of the production source (e.g., in the immediate neighborhood). Specifically, the project addressed the concept of a peer-to-peer market, where owners of solar panels can sell their solar energy production to households in the neighborhood. The idea is to use market mechanism to incentivize adaption of energy consumption to the supply situation in the neighborhood. In combination with automated agents that trade on the market and control selected devices, the market mechanisms drive balancing of supply and demand in the local grid [8].

Complementary to the market mechanism, the PeerEnergyCloud project addressed technologies and concepts for cloud-based infrastructures that integrate smart-home data streams, analyze the data, and provision services on top. The cloud concept is followed for three reasons:

- Hosting the solution as a cloud service reduces the burden for often small sized IT-departments of local grid operators in Germany.
- The cloud allows to leverage the economics of scale when smart home technology becomes more wide spread.
- It is envisioned to provide additional cloud services on top of the smart home infrastructure, sensor data, and analytics capabilities that were investigated in the PeerEnergyCloud project.

Figure 1 gives a high level conceptual overview of the system architecture that was piloted in PeerEnergyCloud and is the target environment for our work. Further details are given in [17].

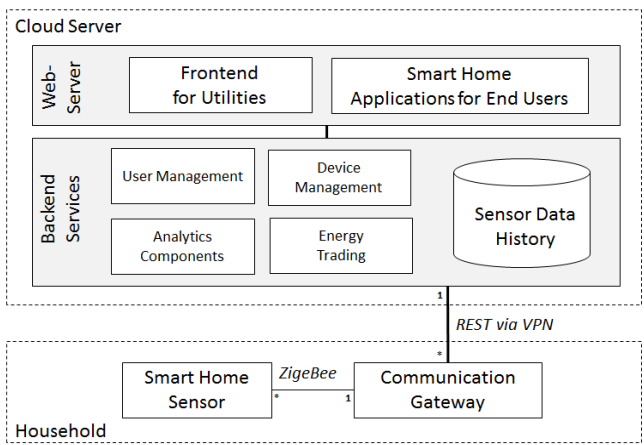

Fig. 1: High level architecture of the system deployment in the PeerEnergyCloud project

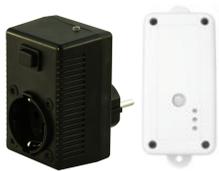

Fig. 2: Sensor devices used for data collection

## 3 LIVED Datasets

The LIVED (**L**ong Dev**i**ce Le**v**el **E**nergy **D**ata) dataset is a subset of the data that was used for the research presented in this paper. The data set has been published as part of the HOBBIT project[1] and is freely available for research purposes[2]. HOBBIT is a research project funded by the European Commision as part of the Horizon 2020 funding framework.

### 3.1 LIVED Sensors

The data set includes both energy readings and additional sensor data. The energy readings have been collected using a Smart Energy Meter from Pikkerton[3], a device to which we refer to as *smart plug*. Occupancy detection, temperature and light conditions were recorded using a *multi-sensor*, see Figure 2.

The smart plugs measure power, frequency, on-off state, voltage, current and electrical work. Multi-sensors provide motion, brightness and temperature readings as well as information about the battery state and voltage. Table 1 shows the available measurement types.

[1] www.project-hobbit.eu
[2] http://lived.agtinternational.com
[3] www.pikkerton.eu

Table 1: Smart plug measurements, reproduced [22]

| Measurement type | Unit | Data type | Resolution |
|---|---|---|---|
| Power | Watt | Float | 1 Watt |
| Frequency | Hertz | Float | 0.0625Hz |
| On state | n/a | ON—OFF | - |
| Voltage | Volt | Float | 1 Volt |
| Current | Ampere | Float | 1 ma |
| Work | kWh | Float | 1Wh |

Table 2: Multisensor measurements, reproduced [22]

| Measurement type | Unit | Data type | Resolution |
|---|---|---|---|
| Motion | n/a | Float | 1-255 |
| Battery Stater | n/a | Text | LOW/OK |
| Brightness | Lumen | Float | 0-2000 Lux |
| Battery Voltage | Volt | Float | 0.01 V |
| Temperature | Celsius | Float | 0.1 C |

| HOUSE | Plugs | Sensors | Total | Measurements |
|---|---|---|---|---|
| House 1 | 6 | 3 | 9 | 522,269,428 |
| House 2 | 7 | 4 | 11 | 636,928,123 |
| House 3 | 7 | 4 | 11 | 534,912,560 |
| House 4 | 7 | 3 | 10 | 159,202,670 |
| House 5 | 7 | 4 | 11 | 648,517,430 |
| House 6 | 7 | 0 | 7 | 18,822,263 |
| Total | 41 | 18 | 59 | 2,520,652,474 |

Table 3: Measurements per houshold available in the LIVED dataset. Reproduced from [22]

### 3.2 Dataset description

The published version of the LIVED data set contains about 2.5 billion measurements collected from 6 households over a period of two years. A typical deployment of sensor devices consists of seven smart plugs and four multi-sensors. The breakdown of devices and measurements per household is shown in Table 3. Leonardi et al. have chracterized the data in more detail [22]. The data is available as compressed CSV files that contain anonymized sensor readings. The format is as follows:

- **TS (timestamp)** − a timestamp representing the time when the measurement was taken
- **Type** − a string representing the measurement type, see Table 1
- **Value** − a string representing the measurement value
- **Unit** − a string describing the unit of measurement
- **Houseid** − an integer representing the house
- **Mac** − MAC address of the sensor. The original MAC address is replaced with an artificially created for privacy reasons
- **Sensor_id** − string identifying the sensor. The sensor identifier contains the houseid concatenated by a unique sensor identifier.

- **Year** – year identifying the year of the measurement
- **Month** – month identifying the month of the measurement

The sensor id uniquely identifies a sensor within the whole data set. Each sensor is either a smart plug or a multi-sensor. The sensor type relates to one entry in Table 1. The year and month fields provide redundant data that is already contained in the timestamp field.

Energy readings are typically configured to sample and transmit their data every two seconds. Sensor devices send their measurements via ZigBee to a home gateway that is connected to the backend infrastructure via the Internet. The timestamp in the data is created when the gateway received the measurements from the sensing devices. The differences of the timestamps between two consecutive measurements may therefore vary considerably in the data. A sample of continuous work readings is shown in Table 4.

## 4 Prediction with the LIVED Dataset

In our experiment, we evaluate timeseries prediction on the full dataset collected within the PeerEnergyCloud project. Parts of the dataset that served for the evaluation became the LIVED dataset. We evaluated multiple prediction algorithms with the goal to beat the benchmark Persistence [30]. Persistence is the most simple approach to prediction, i.e., assumption is that the future consumption is equal to the current consumption. We challenge the benchmark with the algorithms included in the Python package Scikit-learn [26].

The obtained raw dataset contained 36 households. Not all households had good data quality. Typical issues are sensor outages, too few sensors deployed or less than 4 months of uninterrupted data available. The criteria for selecting a household were the following: At least 4 month of data with good data quality. If in the dataset no uninterrupted 4 month period could be found, we took a larger chunk and skipped the days which had data quality issues. Table 5 shows an overview of the 18 selected households, how many days in total were taken, how many days were skipped because of temporary outages and how many days were used to evaluate the prediction algorithms.

### 4.1 Timeseries prediction

In time series prediction, we want to predict, based on the information available at the current point in time, the prediction target which is in future. The features for the prediction are extracted from a time-based window, the target is the aggregated consumption of all devices within a household during the next window. The training data for prediction is the historic window, which is a time-based sliding window. Figure 3 shows the sliding historic window and the forecast in the future. As new data comes in, historic data gets removed from the training data. We have decided to limit the amount of history used for prediction because the prediction results were better and the computational requirements for prediction were reduced. An explanation why short history works better might be that too old data does not reflect the more recent patterns of consumption behavior introduced by different work schedules of the residents, weather, holiday, new electric devices, seasonal changes or other behavior changes.

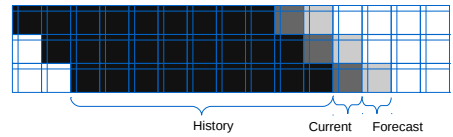

Fig. 3: Window history and forecast

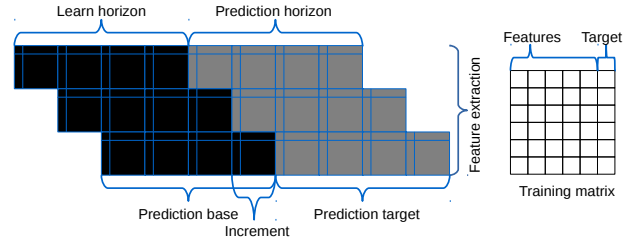

Fig. 4: Micro windowing

To extract more training data from limited history, we used an approach that we call micro-windowing, see Figure 4. The historic window is splitted into sliding windows and for each increment, the features from the prediction base and prediction target are extracted. This is then converted to a matrix of features and used to train the machine learning kernel.

Figure 4 shows an example of a parametized micro-window where every 15 minute increment, the features of the last 60 minutes are extracted to predict the next 60 minutes. The features and prediction targets are then converted to the training matrix and used to train the machine learning classifier.

### 4.2 Error metrics and evaluation

Household energy usage, especially of non-controllable loads, reflect human behavior. The behavior can change over time. Changes of working time, change of jobs, changes based on weather are reflected in consumption traces of the individual appliances. Hence, it is not reasonable to validate using standard k-fold cross-validation. With this approach it would be possible that future values are in the training set and the validation is done on historic data. This approach is clearly not viable for the evaluation of timeseries prediction. A better approach is called timeseries cross-validation [19]. In this approach it is allowed to use as much history as possible to predict future values. We use a sliding window which restricts history to 14 days. After training based on the history we predict one step ahead. After that we compare the prediction with the actual values and calculate the prediction error.

| TS | Type | Value | Unit | Houseid | Mac | Sensor_id | Year | Month |
|---|---|---|---|---|---|---|---|---|
| 2013-10-01 00:09:39 | WORK | 74.973 | kWh | 1 | 00:00:00:00:00:00:11 | 1:98 | 2013 | 10 |
| 2013-10-01 00:09:42 | WORK | 74.973 | kWh | 1 | 00:00:00:00:00:00:11 | 1:98 | 2013 | 10 |
| 2015-02-01 00:00:00 | LOAD | 49 | Watt | 2 | 00:00:00:00:00:00:45 | 2:201 | 2015 | 2 |
| 2014-03-01 00:00:36 | LOAD | 23 | Watt | 6 | 00:00:00:00:00:00:32 | 6:1 | 2014 | 3 |
| ... | | | | | | | | |

Table 4: A sample from the LIVED data set showing work and load measurements

| id | 237 | 244 | 2421 | 2422 | 238 | 212 | 2221 | 241 | 211 | 2401 | 2402 | 226 | 253 | 228 | 221 | 239 | 248 | 215 |
|---|---|---|---|---|---|---|---|---|---|---|---|---|---|---|---|---|---|---|
| total | 181 | 157 | 168 | 183 | 138 | 144 | 146 | 183 | 259 | 120 | 123 | 223 | 141 | 292 | 184 | 172 | 139 | 144 |
| skipped | 4 | 5 | 17 | 15 | 4 | 14 | 6 | 5 | 8 | 2 | 7 | 24 | 6 | 60 | 15 | 24 | 5 | 15 |
| used | 177 | 152 | 151 | 168 | 134 | 130 | 140 | 178 | 251 | 118 | 116 | 199 | 135 | 232 | 169 | 148 | 134 | 129 |

Table 5: Households and ranges

Prediction error is often reported as MAPE (**M**ean **A**bsolute **P**ercentage **E**rror), see Equation 1. The main problem of MAPE is that a division by zero occures when the actual values are zero. In some households this can happen at night or when no sensor is deployed at constantly running devices. Hence we also report the NRMSE (**N**ormalized **R**oot **M**ean **S**quare **E**rror), see Equation 1.

$$\text{MAPE} = \frac{1}{n} \sum_{t=1}^{n} \left| \frac{A_t - F_t}{A_t} \right|,$$
$$\text{NRMSE} = \frac{\sqrt{\frac{\sum_{t=1}(A_t - F_t)^2}{n}}}{\sqrt{\frac{\sum_{t=1} A_t^2}{n}}} \tag{1}$$

### 4.3 Feature extraction

Table 6 gives an overview of the evaluated features. There are two main categories: Features which are calculated over all input data from all devices and per device features.

We calculate the consumption within a window from the measured watts. The plug meters send every few seconds the current load and the total work. The resolution of absolute work values is too low for short-term prediction. We take the average load of each device within a window and summed them up to get the total load. Momentum and WilliamsR (%R) are technical indicators used to interpret financial market movements [20]. The main usage within forecasting is to give the prediction algorithm an idea of how the load developed over the last window. %R (Equation 2) shows the current value in relation to the high and low within a window $w$ [20].

| Household features | |
|---|---|
| summed | average load of each device summed |
| hour | hour of the day, 0-24 |
| wday | Weekday extracted from the date, 0-6 |
| **Device specific features** | |
| min | minimum load in watt |
| max | maximum load in watt |
| mean | average load in watt |
| variance | variance of load in watt |
| stddev | standard deviation of watt measures |
| skewness | skewness of load measurements |
| kurtosis | kurtosis of load measurements |
| moum | Monmentum, financial indicator derived from watt measures |
| willr | WilliamsR, financial indicator derived from watt measures |
| consum | Consumption in kWh over timespan |
| first | first load in timeslot |
| last | last load in timeslot |
| state | last measurement, 1 if more than 0 watt, 0 if less or equal |

Table 6: Overview feature extraction

$$\%R_w = \frac{max_w - m_{wend}}{max_w - min_w} \times -100$$
$$M_w = m_{wend} - m_1 \tag{2}$$

Momentum (Equation 2) is an indicator which shows how quickly the prices are rising or falling [20]. In our case it shows the increase of the energy usage within a time-window. The positive or negative numbers help the prediction algorithm to learn if a device was switched on or energy usage increased or decreased.

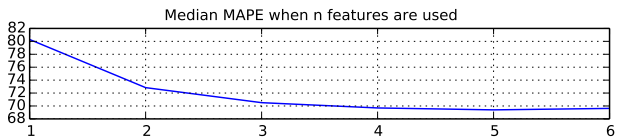

Fig. 5: Median error with more features

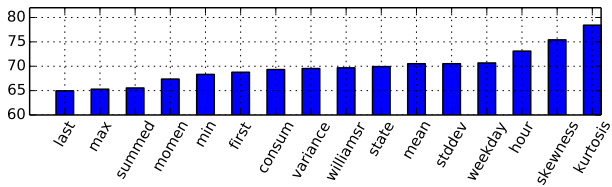

Fig. 6: Median feature performance

### 4.4 Feature selection

To determine the best combination of features, we evaluated the first 6 permutations (Equation 3) of feature combinations on 4 representative households. The training period was set to two month and two month of data were used for evaluation. No relearning was done, so only the first two month were used as a training set. The prediction errors were higher since no continuous retraining was used. This was a compromise to come to a computationally feasible set.

The main conclusions were: When more features are used, the prediction gets only better up to a certain point. This might be due to over fitting on certain features and limited history. Figure 5 shows how the median of the MAPE decreases the more features are added. Using more than four features does not result in a lower MAPE or it can even go up. Certain feature combinations are very good at some households but do not work in average. Figure 6 shows the median MAPE when the feature was included in the feature vector and used for prediction. The first 6 permutations, see Equation 3, were evaluated and the results shows that when the feature *last* was used in combination with other features the median MAPE is lower compared to when *Kurtosis* is used.

$$\mathbb{F} = \{summed, hour, ..., last - state\}$$
$$Feature combinations = \mathcal{P}(\mathbb{F}, 6) = \frac{\mathbb{F}!}{(\mathbb{F} - 6)! 6!} \quad (3)$$

Further analysis, see Table 7, showed that although the feature *weekday* seems to be not the best according to Figure 6, in combination with summed it performed

| Feature combination | Stddev | MAPE | Score |
|---|---|---|---|
| wday-moum-willr-first-last | 27.75 | 51.94 | 79.69 |
| wday-moum-consum-first-last-state | 27.15 | 52.57 | 79.73 |
| wday-max-willr-last-state | 29.37 | 50.46 | 79.82 |
| wday-moum-last-state | 26.48 | 53.53 | 80.00 |
| wday-moum-consum-first-last | 27.77 | 52.30 | 80.07 |
| wday-min-williamsr-last-state | 28.12 | 53.72 | 81.84 |
| sum-wday-stddev-willr-consum-last | 30.03 | 51.99 | 82.02 |
| wday-max-will-r-consum-last-state | 30.65 | 51.55 | 82.20 |
| wday-min-max | 30.90 | 51.37 | 82.26 |
| ... | | | |

Table 7: Standard deviation (Stddev) of MAPE and average MAPE

quite well. Table 7 shows the top feature combination ranked by score.

$$h = \{1, 2, ..., end\}$$
$$\mathbb{H} = \{209, 237, ..., 215\}$$
$$AvgMAPE = \frac{\sum_{H_{i=0}}^{H_{end}} MAPE_i}{end}$$
$$Stddev = \sqrt{\frac{1}{N-1} \sum_{i=1}^{N} (x_i - \overline{x})^2} \quad (4)$$
$$Score = AvgMAPE + Stddev$$

The Standard Deviation (Stddev) shows how stable the feature combination is at multiple households. The lower the Stddev is, the more useful the feature combination is at multiple houholds. The lower the average MAPE is, the better are the predictions. There were several feature combinations which have led to a very low AvgMAPE (Averaged MAPE of all households) but a very high Stddev and vice versa. Hence we have chosen to weight both parts in a overall measure called Score, see Table 7.

The more features are used, the higher the standard deviation of the error. This is likely a result of over fitting with a specific feature combination. A surprise for us was that the hour of the day did not show a better performance. The main reason for this is that the dataset mainly has instantaneous devices and there is no clear indication that the usage depends on the time but is more correlated with the behavior of the time slot before. However, predictions tend to have less Stddev when the feature summed was included. Based on the score, detailed analysis of outliers and several runs of some other combinations we have decided to evaluate two different feature combinations for in-depth evaluation, see Table 8.

| Name | Short | Feature combinations |
|---|---|---|
| Complex | cpx | summed, weekday, lastvalue, max-value, williamsrvalue |
| Minimal | min | summed, weekday |

Table 8: Feature combinations

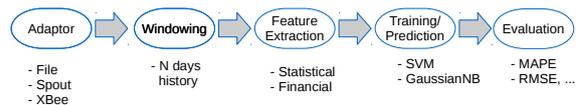

Fig. 7: Iterator chain

| Name | Regression | Classification |
|---|---|---|
| Support Vector Machine | ✓ | ✓ |
| Gaussian Naive Bayes | | ✓ |
| Decision Tree | ✓ | ✓ |

### 4.5 Prediction algorithms

We used several classification and regression algorithm of the machine-learning package scikit-Learn [26] and evaluated both the computational requirements and the prediction errors when used for forecasting household load.

A classifier can predict discrete classes, since load is a continuous value we used K-Means to discretize. The rational behind that is that there is only a limited amount of devices and not all will be used at the same time. Hence there is also a limited number of load combinations. Most devices have on-off patterns which means two different kind of loads. We assumed 8 clusters. The classifier predicts then one of these clusters, the centroid of the cluster was taken as the prediction.

## 5 System Architecture

In this section we present the architecture of the analytic component shown in Figure 1. The goal we want to reach is that we can use a single codebase to predict continuously streaming data and to analyze historical data.

Continuous timeseries prediction requires the extracted features and targets from a limited history for training. The training history is limited to several weeks. The current features are used to predict in the future. Our implementation is based on iterators and by chaining them together we build a prediction pipeline. The iterator exposes one function called *next* which advances to the next event. One iterator can pull from another iterator and this way it is possible to build pipelines. We built the whole machine learning pipeline in terms of these iterators and chained them together, see Figure 7. Depending on if the pipeline is used in batch analytics or within streaming, the adaptor emits either a log of events or directly events from the stream. The windowing iterator emits whole windows of events every increment. The next iterator pulls these windows and extracts the features and further emits the features to the prediction iterator. This iterator again pulls from the feature extraction iterator and predicts in the stream. The last part of the prediction pipeline can be either be a analytics component which calculates the prediction error or other components like an agent participating in the market or a model predictive control algorithm. By chaining these iterators we can reuse the complex parts of the implementation by wiring the chain differently.

### 5.1 Implementation

We create an adaptor for Apache Storm [1] to predict in the stream. Apache Storm provides two abstractions, Bolts and Spouts. Spouts are used for event sources and Bolts for processing logic. The adaptor bridges the Bolt API to an iterator. The stages of the prediction pipeline can be expressed in multiple bolts to increase parallelism. Since multiple households are predicted at the same time, we used the household identifier as a partition key and instantiated one prediction pipeline per household. As shown in Figure 7, all stages expose their result as an iterator. Consequently each stage could be run as a separate Bolt and can be distributed across multiple nodes. Figure 8 shows an overview over the implementation. The gateways in the households send data over a secure channel to the prediction pipeline. The Storm topology contains a Spout and a prediction Bolt per household. Finally the results of the pipeline are aggregated and analyzed.

## 6 Results

In this section we show the results regarding the prediction accuracy, stability of the prediction error and the computational impact.

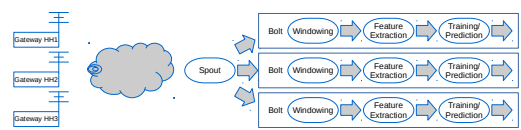

Fig. 8: Parallel prediction pipelines

### 6.1 Prediction error

First we discuss the results of the prediction algorithms, then we put our results in perspective to related work. Table 9 shows the Median MAPE of the different methods at the households. It is clearly possible to beat the benchmark. Until the 60 minute ahead forcast classifiers performed better then regressors. Until 6 hours ahead forcast, regression based method performed slightly better then classification methods. The error measure MAPE has drawbacks in case the actual value is zero. Some of the households have quite often zero consumption during nights, thus we can use NRMSE to see if the specific household is affected. We can see that regression preforms clearly better with this metric. However, in general we can see that also using NRMSE, the regression based support vector machine using the complex or minimal feature set works best for this kind of prediction. The worst result was achieved using Gaussian Naive Bayes with the complex feature set.

Another machine learning case study [10], in which whole house consumption was predicted one hour ahead, reported a MAPE of $\approx 20\%$ using Support Vector Regression. Also [11] shows that the prediction error of a whole household is around $\approx 20\%$ using a live error correction technique. However in our dataset there is a limited number of devices, hence the prediction error is higher at $\approx 40\%$. In Household Electricity Demand Forecasting - Benchmarking State-of-the-Art Methods [30] multiple prediction methods were evaluated. The hour ahead forecast with Persistence was 60% which is roughly comparable with the performance of Persistence in our evaluation. As our dataset is derived from individual plug meters, it is a lot spikier than a whole household where heat-pumps, freezers or other devices present a base load. In our case we can see that in the small, SVRs can beat Persistence. The best 30 minute ahead forecast in their evaluation was using Neural Networks with 49%. This is comparable with our results, which was $\approx 48\%$ using Support Vector Classification.

To illustrate why prediction of accumulated devices is different than forecasting whole house consumption or even prediction at higher aggregations, we show a trace of two days, see Figure 9. The x-axis shows the total amount of energy used. The blue line denotes the actual values, the green line what the algorithm has forecasted. We can see very spiky consumption patterns. Nevertheless there were some periods where the prediction was really good (12:00 - 18:00 on Aug. 8th). However the spikes between 00:00 and 12:00 on Aug. 8th were not predicted at all. Furthermore the spike between 20:00 - 24:00 on Aug. 9th were predicted quit well. Overall the graph shows how random the aggre-gated consumption of multiple devices is and that there are periods with nearly no prediction error and periods where the prediction is off.

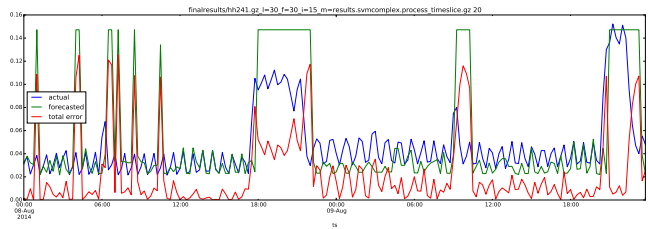

Fig. 9: Exemplary prediction

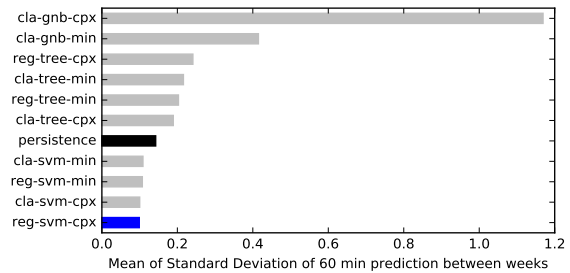

Fig. 10: Stability of the prediction error

### 6.2 Prediction error stability

As shown in Figure 9, the load is very spiky. This motivates the analysis of the prediction error stability since it is not favorable that the prediction works a few days well and then not at all. In case we want to compensate energy usage caused by instantenious human behavior with batteries, we need to be able to quantify the stability of the prediction error. Not only should the prediction error be small overall, but also not small for several weeks and then a week with high prediction error.

In the first step we calculate the MAPE per week of each household, then we calculate the standard deviation of the prediction errors of 8 subsequent weeks. This is done for each predictor and feature combination and the mean of the households is taken. Figure 10 shows the mean standard deviation for 1h forecast. We can see that the predictions of support vector regression with the complex feature set are more stable then the benchmark persistence.

### 6.3 Performance evaluation

We evaluated the performance of the machine learning kernel during continuously prediction of the future. All evaluations were conducted on a machine with 4xIntel E7-4850 v3 @ 2.20GHz and 0.5TB RAM.

| MAPE | Classification | | | | | | Benchmark | Regression | | | |
|---|---|---|---|---|---|---|---|---|---|---|---|
| | gnb-cpx | gnb-min | svm-cpx | svm-min | tree-cpx | tree-min | persistence | svm-cpx | svm-min | tree-cpx | tree-min |
| 15 | 739.04 | **61.63** | 82.10 | 68.47 | 83.52 | 113.94 | 91.14 | 79.80 | 75.31 | 66.21 | 109.41 |
| 30 | 801.48 | 68.48 | 60.11 | **47.76** | 97.92 | 132.57 | 71.42 | 65.44 | 56.82 | 76.28 | 127.65 |
| 60 | 808.60 | 79.57 | 39.65 | **38.72** | 86.32 | 120.64 | 73.15 | 44.59 | 39.63 | 95.50 | 124.14 |
| 90 | 473.53 | 113.54 | 37.03 | 41.20 | 98.20 | 116.58 | 83.12 | 39.42 | **36.35** | 104.85 | 115.72 |
| 120 | 336.57 | 121.31 | 36.62 | 40.53 | 101.67 | 117.26 | 84.36 | 39.30 | **36.28** | 111.85 | 117.69 |
| 360 | 201.05 | 88.82 | 47.78 | 63.36 | 102.56 | 123.35 | 112.91 | 46.42 | **46.25** | 102.40 | 127.02 |
| 720 | 104.49 | 82.56 | **46.25** | 69.67 | 76.02 | 92.16 | 98.50 | 51.51 | 56.08 | 78.18 | 95.20 |
| 1440 | 53.26 | 55.92 | 48.18 | 47.04 | 58.37 | 57.00 | 54.89 | **45.48** | 48.17 | 59.06 | 57.78 |

Table 9: Median MAPE from all households

| NRMSE | Classification | | | | | | Benchmark | Regression | | | |
|---|---|---|---|---|---|---|---|---|---|---|---|
| | gnb-cpx | gnb-min | svm-cpx | svm-min | tree-cpx | tree-min | persistence | svm-cpx | svm-min | tree-cpx | tree-min |
| 15 | 220.53 | 85.18 | 91.02 | 90.47 | 80.87 | 92.93 | 78.88 | 91.54 | 91.15 | **71.21** | 94.22 |
| 30 | 229.93 | 95.40 | 89.51 | 88.89 | 88.50 | 106.30 | 85.64 | 89.03 | 88.79 | **84.51** | 104.13 |
| 60 | 269.17 | 106.83 | 87.73 | **86.66** | 92.33 | 104.81 | 94.45 | 86.34 | 86.11 | 90.63 | 106.44 |
| 90 | 221.17 | 120.05 | 86.32 | 86.14 | 93.73 | 100.13 | 93.46 | 84.20 | **83.88** | 99.71 | 101.46 |
| 120 | 164.72 | 108.93 | 86.05 | 83.10 | 93.31 | 97.40 | 92.14 | 82.12 | **81.82** | 94.94 | 98.58 |
| 360 | 121.95 | 95.73 | 76.60 | 81.16 | 82.90 | 92.28 | 85.22 | 70.60 | **70.06** | 88.83 | 91.28 |
| 720 | 79.49 | 78.63 | 65.80 | 65.62 | 68.47 | 73.52 | 76.42 | 57.66 | **56.16** | 70.63 | 75.03 |
| 1440 | 51.50 | 50.38 | 54.39 | 51.34 | 51.61 | 53.81 | 51.01 | 44.16 | **42.17** | 56.05 | 53.44 |

Table 10: Median NRMSE from all households

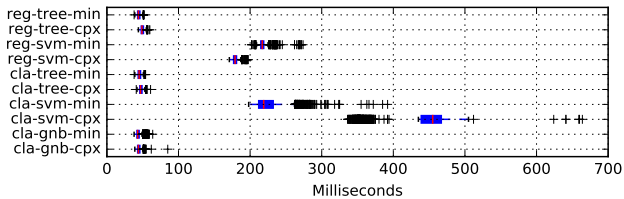

Fig. 11: Predictorperformance

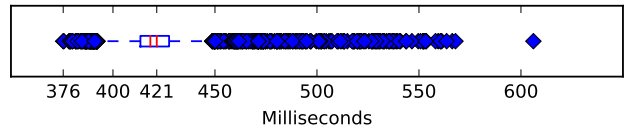

Fig. 12: Latency of prediction pipeline

First, we evaluated the performance of the predictors with different feature combinations. Figure 11 shows Boxplots of how many milliseconds a single prediction takes. One trend is the more features are used, the longer the prediction takes. We did not include the benchmark persistence since it is only limited by the performance of feature extraction.

In the second evaluation we used Apache Storm and measure the performance of the prediction pipeline. We measure the time from the timestamp the data was handed of till the prediction arrives at the next spout. This included windowing, feature extraction and prediction which is implemented as a single Storm Bolt. We use as the machine learning kernel support vector regression with the complex feature set. We use 15 days for training and predicted every minute based on the last 15 minutes the next 15 minutes. Figure 12 shows the latency distribution with the mean and median latency around $420ms$. This means, compared to the sole latency of the predictors, $\approx 200ms$ is used for windowing and feature extraction and other overheads.

# 7 Related work

In this section we list the related work in the area of prediction in the stream and related work in the area of household energy prediction.

**Predictability of Energy Use in Homes** [21] is a study about the predictability at different granularities, whole household and single appliances. The dataset the authors use was derived from 7 sites and includes individual appliances (6-21) and whole house energy consumption. The scenario they consider is different, hence they also consider different error metrics. The primary metric they consider is accuracy. The tolerance when a prediction is accurate is $\pm 5\%$ of the maximum power draw ever observed of the appliance. In our dataset, the power draw of individual devices was very high. A dishwasher could use $\approx 1,700\ Watts$ when heating the water and then continue at $\approx 150\ Watts$. In that case the tolerance would be $85\ Watts$ which would already the normal consumption of a flatscreen. In a scenario, where we cover mainly non-controllable loads and try to predict the aggregate, we have to consider measures which take into account the aggregate consumption.

**Predicting future hourly residential electrical consumption: A machine learning case study** [10] evaluates multiple machine learning algorithms with 10-folds cross validation. This work predicts whole houses including HVAC systems or cooling devices. The houses are equipped with approximately 140 different sensors that collect data every 15 min. The occupancy is simulated with automated controls which open and close fridges and so on. Dynamic human behavior was removed by simulating occupancy. There are still some dynamic changes by prototype equipment tests, thermostat set point changes and so on.

Since our dataset mainly includes instantaneous loads like entertainment or lights and is also derived from real world households with all the dynamic human behavior, the paper provides an interesting base for discussion about the effect of human behavior changes on the prediction quality. Since we are using timeseries prediction, we cannot use 10-fold cross validation.

**Household Electricity Demand Forecasting - Benchmarking State-of-the-Art Methods** [30] shows an evaluation of the prediction error of multiple forecasting algorithms. The main conclusion is that the most simple forecasting algorithm, the consumption in the last timeslot will be consumption of the next timeslot, is a very hard to beat benchmark. The sample size used in this paper was two households with several month of data. In the result section we discuss our results in perspective the paper and we also use the same benchmark.

**One hour ahead prediction for households at different aggregation levels** [28] shows the effect on the prediction error when multiple households are aggregated. In our work we show also the effects of aggregation but on a smaller scale, several devices within a household aggregated to the total consumption.

**Short-term smart learning electrical load prediction algorithm for home energy management systems** [11] shows a method which creates first a day-ahead forecast and then incrementally adapts the prediction as new data comes in. Altough the scenario they imagined was similar to ours, the dataset is derived from a whole household. In whole houses their are typically water pumps, fridges and freezers running having a constant load. When the demand is derived from instantaneous devices only the patterns are spikier and less predictable. The error is measured using MAPE, which makes the results comparable to our work.

**Energy Forecasting for Event Venues: Big Data and Prediction Accuracy** [14] shows how big data and prediction can be brought together to predict daily, hourly and 15 minute forecasts. The main difference in the approach is that they take hoping windows as input. In our approach we have chosen micro-windowing and a limited amount of history.

## 8 Conclusions and future work

We presented an architecture for prediction in the stream, an approach called micro-windowing to extract more training data from a limited timespan and an evaluation of prediction accuracy and computational requirements of our approach. On top of that, we analyse a range of prediction mechanisms and feature sets for household specific load forecasts and provide insights into their performance. We thereby increase the overall understanding of the short-term predictability of individual household loads and the influencing factors. Additionally, parts of the dataset the evaluation is evaluated on, the LIVED dataset is opened and available for further research.

We see future work in the interaction of prediction and the control of devices. A scenario could be that a short high energy prediction and low solar in-feed adjusts the thermostat of the freezer temporary without effecting the comfort.

In our approach we retrain fully for each prediction. Altought this takes only several milliseconds, the computational requirements could be reduced further. Incremental and decremental learning [24] is an interesting approach to limit the training horizon on a specific timeframe. With this approach it would be possible to incrementally learn and forget and continously retraining the machine learning kernel would not be necessary.

In the dataset additional measurements of multisensors are available. We haven't considered them for prediction since not in all of 18 households they were deployed accurately. We think that movement sensors could reduce the prediction error. Another way to reduce the prediction error would be to take the class of device into account. Certain devices, e.g., washing-machines or dish-washers, have once switched on a known pattern over the next several hours. If this features would be taken into account, the prediction error could be reduced.

## 9 Acknowledgements

This work has been supported in part by the Alexander von Humboldt foundation. The work presented in this paper has partly been funded by the H2020 project HOBBIT under the grant agreement number 688227. This work has been supported in part by the German Federal Ministry of Economics and Technology, project PeerEnergyCloud which is part of the Trusted Cloud Program.